\documentclass[sigconf]{acmart}

\setcopyright{none}  
\renewcommand\footnotetextcopyrightpermission[1]{}  

\acmConference[Augmented Educators and AI(CHI 2025 Workshop)]{Augmented Educators and AI: Shaping the Future of Human-AI Collaboration in Learning on CHI 2025 Workshop}{April 26,2025}{Yokohama, JAPAN}

\begin{document}

\title{“Clicking some of the silly options”: Exploring Player Motivation
in Static and Dynamic Educational Interactive Narratives}

%Dany - Sam - Alex - Shi - Michael - Noah - Eddie

\author{Daeun Hwang}
\orcid{0000-0002-7587-1674}
\affiliation{%
  \institution{University of California, Santa Cruz}
  \city{Santa Cruz}
  \country{USA}
}

\author{Samuel Shields}
\affiliation{%
  \institution{University of California, Santa Cruz}
  \city{Santa Cruz}
  \country{USA}
}

\author{Alex Calderwood}
\affiliation{%
  \institution{University of California, Santa Cruz}
  \city{Santa Cruz}
  \country{USA}
}

\author{Shi Johnson-Bey}
\affiliation{%
  \institution{University of California, Santa Cruz}
  \city{Santa Cruz}
  \country{USA}
}

  \author{Michael Mateas}
\affiliation{%
  \institution{University of California, Santa Cruz}
  \city{Santa Cruz}
  \country{USA}
}

  \author{Noah Wardrip-Fruin}
\affiliation{%
  \institution{University of California, Santa Cruz}
  \city{Santa Cruz}
  \country{USA}
}

  \author{Edward F. Melcer}
\affiliation{%
  \institution{University of California, Santa Cruz}
  \city{Santa Cruz}
  \country{USA}
}

%%adds a footer with the author infor to each page. 
\renewcommand{\shortauthors}{Hwang, D. et al.}

\begin{abstract}
Motivation is an important factor underlying successful learning. Previous research has demonstrated the positive effects that static interactive narrative games can have on motivation. Concurrently, advances in AI have made dynamic and adaptive approaches to interactive narrative increasingly accessible. However, limited work has explored the impact that dynamic narratives can have on learner motivation. In this paper, we compare two versions of Academical, a choice-based educational interactive narrative game about research ethics. One version employs a traditional hand-authored branching plot (i.e., static narrative) while the other dynamically sequences plots during play (i.e., dynamic narrative). Results highlight the importance of responsive content and a variety of choices for player engagement, while also illustrating the challenge of balancing pedagogical goals with the dynamic aspects of narrative. We also discuss design implications that arise from these findings. Ultimately, this work provides initial steps to illuminate the emerging potential of AI-driven dynamic narrative in educational games.

%Previous research highlights the positive effects of static interactive narratives in educational games, suggesting high learning potential. Concurrently, advances in narrative AI have made dynamic and adaptive approaches to interactive narrative increasingly accessible. However, limited work has explored the impact that dynamic narratives can have on player motivation. Here we compare two versions of Academical, a choice-based educational game about research ethics. One version employs a traditional hand-authored branching plot (i.e., static narrative) while the other dynamically sequences plots during play (i.e., dynamic narrative). Results highlight the importance of responsive design and the variety of choices for player engagement, and suggest that the dynamic narrative resulted in more positive player autonomy experiences overall. Ultimately, our findings illuminate the emerging potential of dynamic narrative in educational games and provide design implications.  
\end{abstract}

\keywords{Interactive Narrative, Dynamic Narrative, Static Narrative, Educational Games, Responsible Conduct of Research}

\maketitle

%%Adds a CC BY 4.0 license note and workshop information the first page. Do not fix it.
{\small
\noindent ©  This paper was adapted for the \textit{CHI 2025 Workshop on Augmented Educators and AI: Shaping the Future of Human and AI Cooperation in Learning},
held in Yokohama, Japan on April 26, 2025. This work is licensed under the Creative Commons Attribution 4.0 International License (CC BY 4.0).
}

\section{Introduction}
It is challenging to find ways to cultivate an understanding of research ethics and motivation to care about them \cite{schmaling2009, kalichman2013, kalichman2014, bouville2008}. This challenge has been addressed in a variety of ways, ranging from traditional expository prose to live-action role-playing of ethical dilemmas \cite{whitbeck2001, brummel2010, kalichman2013b, seiler2011}. While most training in the area remains textbook-like, the use of narrative and interactive methods has found success in engaging learners \cite{melcer2020getting, salter2016, friedhoff2013}. Our research builds upon the finding that interactive visual novels (VNs) can be more effective than traditional training materials for engaging learners and cultivating knowledge of responsible conduct of research (RCR) \cite{grasse2021academical}. However, while existing work has focused on the efficacy of traditional hand-authored branching plot for educational narratives (i.e., static narratives) \cite{grasse2022using}, little has been done to explore the engagement and motivational possibilities of dynamically sequenced plot during play. Engagement and motivation in particular are important to consider in the design of educational technologies as greater engagement and motivation have been shown to lead to greater knowledge transfer and increased achievement~\cite{hsu2019reexamining,guay2008optimal,bond2020mapping}.
 Dynamic narratives are particularly intriguing in this context as their dynamism (i.e., ability to provide a wide swath of content dynamically based upon player choices) offers potential to further engage and motivate players in a variety of ways~\cite{johnson2024academical}.

In this paper, we explore the potential of dynamic interactive narrative in motivating players, examining their engagement  with respect to 1) elements of the game design, 2) perceptions of the narrative structure presented, and 3) the motivational factors of  Self-Determination Theory~\cite{ryan2006motivational}---i.e., autonomy, competence, and relatedness (ARC)~\cite{oliver2016video}. Specifically, we present results from a comparative study contrasting two versions of an interactive VN for learning RCR. The existing game is a statically authored narrative used in existing RCR education research~\cite{melcer2020getting}, which we refer to as "Academical 1.0". The other is designed to offer a novel dynamic narrative approach to the same content presented in Academical 1.0, which we refer to as "Academical 2.0". Our results highlight both the strengths and challenges of incorporating AI into educational interactive narratives, providing design implications to inform the future creation of dynamic educational narratives (DEN).

%One is a subset of the version compared against traditional training materials in earlier research \cite{melcer2020getting}. We refer to this as “Academical 1.0.” The other is a parallel subset of an RCR-focused visual novel that is designed to offer players an experience with more dynamic content and consequential interactivity — a Dynamic Educational Narrative (DEN). We refer to this as “Academical 2.0.”

%Our contribution is threefold. First, our comparison shows that a DEN approach has substantial promise for learning. While the traditional visual novel (Academical 1.0) was characterized as “simplistic” by participants, the DEN (Academical 2.0) was more successful for player motivation in terms of self-determination theory (SDT). Second, we provide insight into autonomy phenomena for interactive experiences, particularly interactive narratives. While our DEN \textit{system} was designed to aid in the authoring of high-autonomy experiences, the \textit{content} constraints of our comparison made it challenging to support autonomy, resulting in similar autonomy findings between Academical 1.0 and 2.0. Third, we explore design implications for effective DEN approaches based on Self-Determination Theory. Drawing from our findings alongside existing literature, we seek to pinpoint essential elements to improve learning efficacy and motivation. Along with these insights, we also discuss the limitations of our study and future research directions. 

\section{Related Work}
\subsection{Dynamic Narrative Systems}

\begin{figure*}
    \centering
    \includegraphics[width=0.8\linewidth]{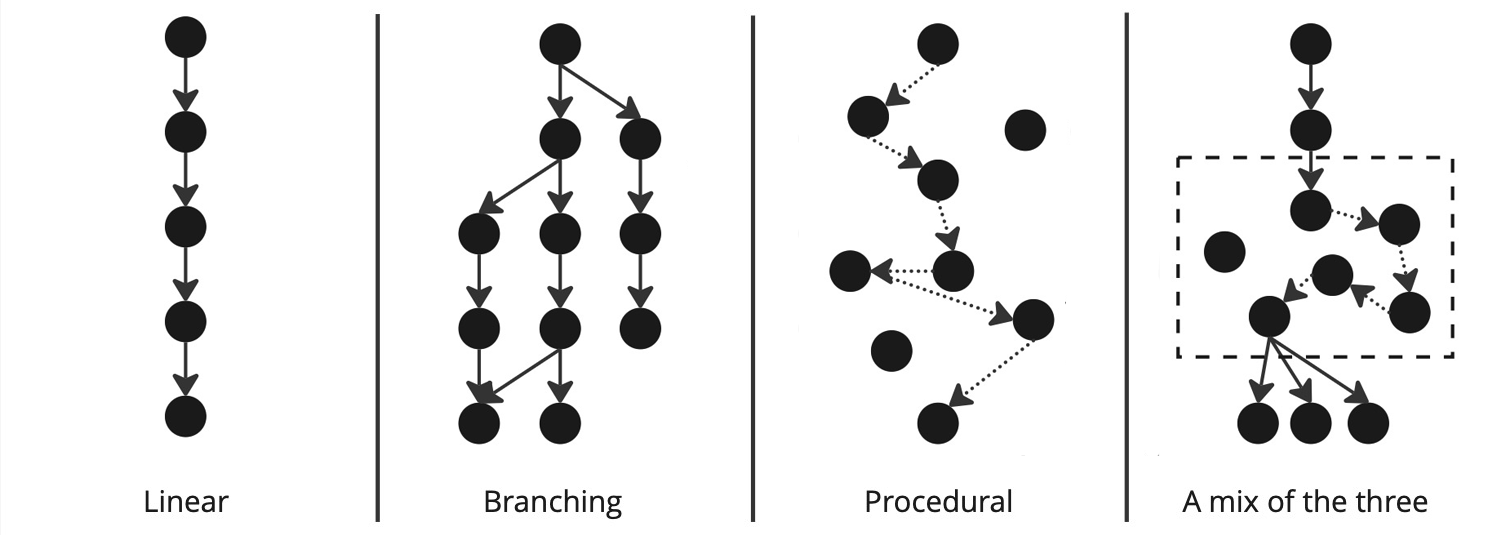}
    \caption{Spectrum of storytelling in games. The circles represent plot points, and the solid arrows are hand-authored connections from one plot point to the next.
    %Linear storytelling moves the player from plot point to plot point without any say. Branching allows the player to move between various predefined plots. Procedural storytelling allows the player to chart their own course through various fragments of story content.
    The dotted arrows represent the plot points connected at runtime by system rules informed by designer goals and the history of player choice.
    }
    \label{fig:plot-structures}
    \Description{A diagram split into three columns. Each column contains circles connected by arrows, representing potential playthroughs of a game's plot. The columns are labeled: "linear", "branching", "procedural", and a "mix of the three"}
\end{figure*}

Storytelling in games exists on a spectrum in terms of dynamism, ranging from linear to branching to procedural (see Figure~\ref{fig:plot-structures}). Linear storytelling moves the player from plot point to plot point without deviating from the core plot. Branching stories use player choices to branch and move the player between predefined branching plot paths. This design approach resembles \textit{Choose Your Own Adventure} (CYOA) novels and is the most common form of interactive narrative in games~\cite{melcer2020teaching}. Lastly, procedural storytelling breaks story content into loosely connected chunks, commonly called ``storylets,'' that players can explore in any order~\cite{kreminski2018storylets}. Computationally, it is the most systems-heavy of the three, but it provides players more freedom in shaping the story plot they experience \cite{short2019storylets}.

\subsection{Narratives and Learning}
VNs are commonly defined as a genre of digital game with a narrative focus in which players shape the story or progression of the game through their interactions \cite{Camingue2021}. VNs are a popular genre for educational games~\cite{Camingue2020} covering topics ranging from second language acquisition~\cite{Faizal2016,Gabriel2018,Amalo2017,Putri2021,Agusalim2012}, literacy~\cite{Lai2021,Sullivan2014}, math~\cite{Nugroho2018,Muntean2018}, awareness~\cite{Andrew2019,Korhonen2017}, training~\cite{Huynh2017}, and health~\cite{Yin2012}. Past studies have shown that utilizing educational VNs is more effective in terms of attitudes, motivation, engagement, skill-development, and learning when compared to non-interactive, traditional text-based approaches~\cite{melcer2020getting,Faizal2016,Gabriel2018,Salazar2013}. However, while many educational interactive narrative games have been created and assessed, they have predominately utilized hand-authored static narrative approaches~\cite{Camingue2020}.

In contrast, dynamic narratives offer the ability to cater learning and story beats to an individual user during their playtime. This allows users to discover pedagogical content at their own pace, as well as the system to serve learning content informed by player navigation~\cite{rowe2010framework}. A notable implementation of this approach comes from the serious game \textit{80days}, which utilizes a dynamic system to adjust narration and learning in response to the ``context and characteristics of individual users or user groups'' \cite{gobel200980days}. The creators note the importance of design-time authoring of a variety of scenes that should be then ordered at runtime according to the learner's traits \cite{koidl2010dynamically}. This style of narrative offers the potential to introduce educational content to the player throughout the course of an experience while simultaneously balancing the presentation of entertainment-focused content to maintain player engagement and motivation~\cite{johnson2024academical}. This in turn is crucial for the effectiveness of educational interactive narrative games~\cite{marsh2011fun}. 

\section{Academical}
%In this section, we present the contrasting static and dynamic narrative designs of Academical 1.0 and Academical 2.0.

\begin{figure*}
    \centering
    \includegraphics[width=\linewidth]{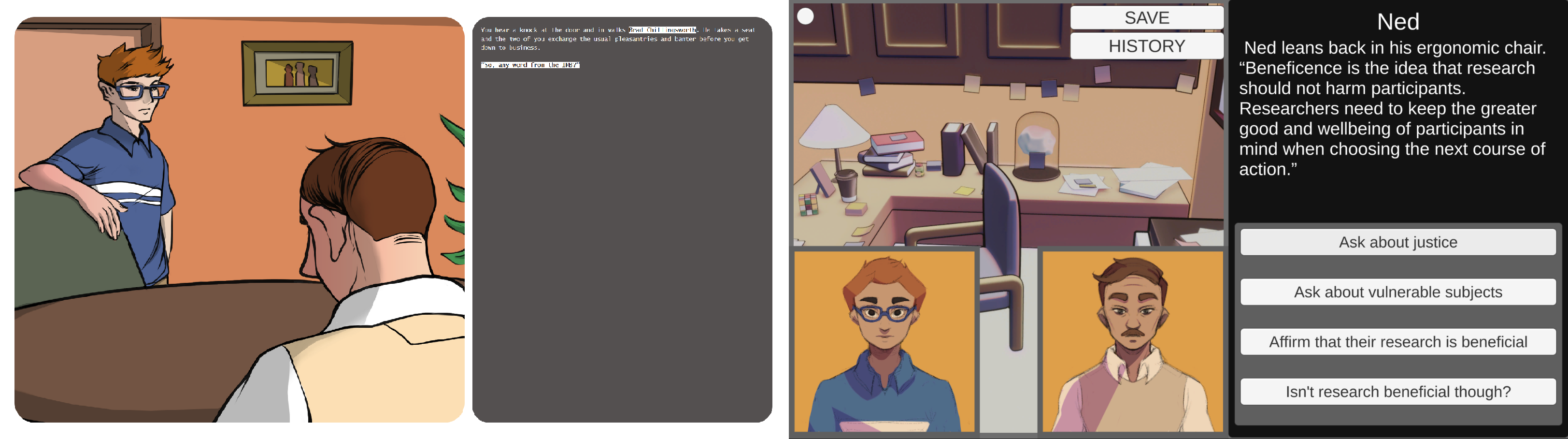}
    \caption{A screenshot of Academical 1.0 (left) next to 2.0 (right).}
    \label{fig:screenshot}
    \Description{Two side-by-side screenshots showing Academical 1.0 next to 2.0. Both feature an image on the left hand side with dialogue and choices on the right hand side.}
\end{figure*}

\subsection{Academical 1.0}
Academical 1.0 is a choice-based interactive narrative game developed to teach research ethics~\cite{melcer2020getting}. The game consists of nine hand-authored static, branching scenarios covering topics in RCR~\cite{brummel2010development}. Furthermore, each of the nine scenarios are designed to show how seemingly obvious answers around questions of research ethics can be complicated by factors such as power dynamics and marginalized identities~\cite{melcer2020teaching}. This in turn leads to a richer understanding of the ethical complications that one can encounter while conducting research. Previous studies of 1.0 demonstrated efficacy in engaging students and teaching three learning outcomes that influence ethical behaviour (knowledge, moral reasoning skills, and attitudes)---particularly when compared to traditional web-based university RCR training modules~\cite{grasse2021academical,grasse2021improving}. Recent work has also explored the effect of motivational components within 1.0, highlighting that while it did lead to need satisfaction for each of autonomy, relatedness, and competence, there was also a fair amount of need frustration that could be addressed by a dynamic approach to the narrative structure~\cite{grasse2022using}.

%Overall, Academical has demonstrated that web-based interactive technologies and games are a promising method for engaging students and teaching them both cognitive and socio-affective learning outcomes simultaneously---which is notoriously difficult to accomplish for standard pedagogy~\cite{}.

\subsection{Academical 2.0}
Academical 1.0 %while covering a large range of scenarios and perspectives, 
featured relatively short, static storylines. Notably, the motivational ARC factors, as well as corresponding literature on narrative games suggest that wider narrative possibilities with high flexibility could provide more autonomy~\cite{wardrip2014towards,harrell2009agency,iten2018choosing}. Autonomy, in turn, could lead to greater competence, higher learning outcomes, and overall engagement for users. Thus, 2.0 aimed to address this by using procedural narrative techniques to enable flexible traversal of sub-topics for pedagogical and role-playing goals. The increased authoring complexity paired with the desire for an "apples-to-apples" comparison between 1.0 and 2.0 led us to design a single scenario for 2.0 translated from one scene in 1.0. Our design approach aimed to increase autonomy in two distinct ways. First, we wanted to offer a wider set of selection choices per dialogue utterance---increasing the number of options from one or two into four, five, or six. Second, we organized pedagogical and drama topics into ``threads'' which fulfilled system-defined learning and narrative goals. These conversation threads would be internally consistent but could be switched between to enable choice in selecting topics throughout the experience. We hypothesized that these approaches should provide autonomy while retaining learning outcomes (e.g., players can't miss critical information). Screenshots from both games are shown in Figure~\ref{fig:screenshot}.

\subsection{Feature differences between version 1.0 and 2.0}

%---------Figure 3--------

\begin{figure}
     \centering
     \includegraphics[width=\linewidth]{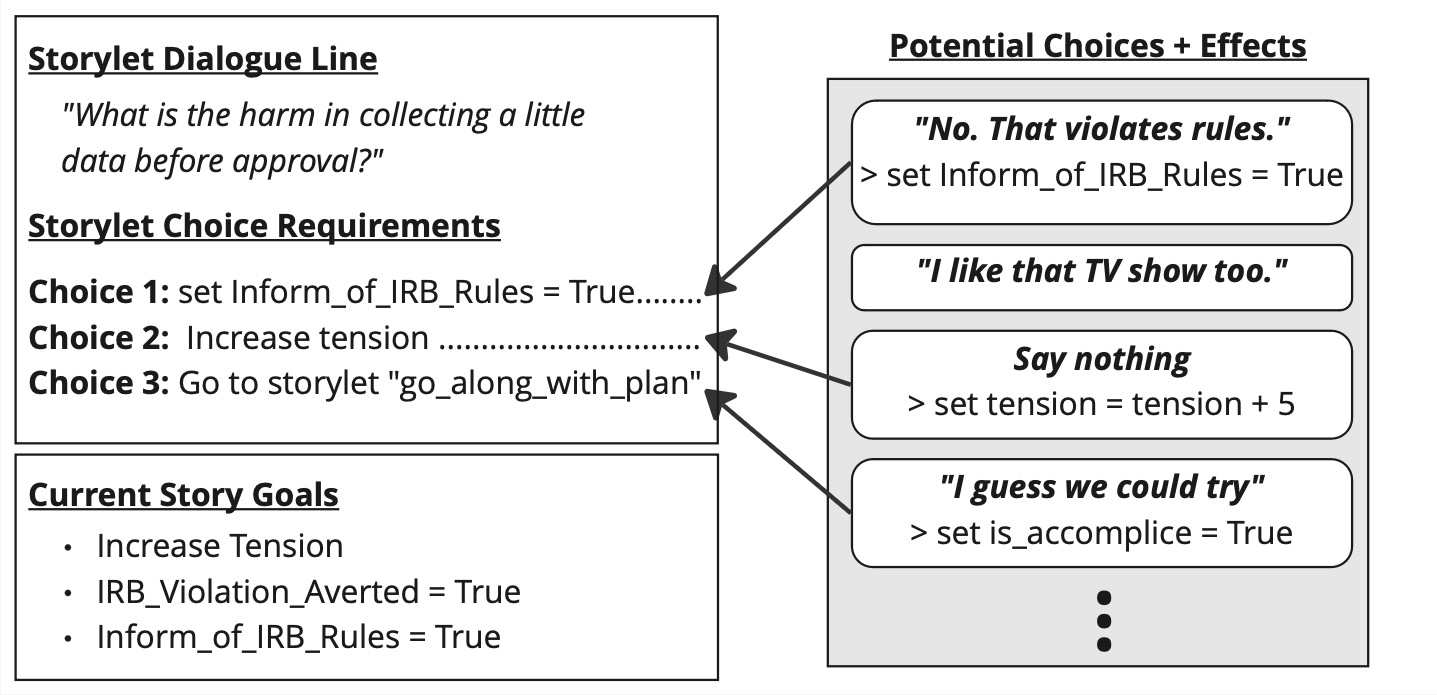}
     \caption{Choice options in StoryAssembler are chosen dynamically from the pool of available storylets. When selecting choices, the system looks at the precondition requirements for each choice (left) and attempts to find a storylet that satisfies that requirement. Additional preference is given to storylets that also accomplish story and/or pedagogical goals.}
     \label{fig:storyassembler-example}
     \Description{A diagram with two boxes. Arrows point from selected storylet choices to their respective requirements in the current storylet.}
\end{figure}

The internal structure of 1.0 is a network graph consisting of passages (nodes with dialogue content) connected by links. Players read the dialogue, click links to transition to different passages, and, based on their link choices, traverse different branches of the predetermined plot. For our purposes, we hypothesized that the presence of pre-authored story branches would limit players’ feelings of autonomy since the narrative is deterministic. Thus, for 2.0, we focused on increasing the level of dynamism and variety in conversational dialogue to increase players' feeling of autonomy. This meant replacing the hand-authored narrative with a structure that dynamically sequenced content at runtime. We accomplished this by recreating a subset of the StoryAssembler architecture~\cite{garbe2019storyassembler} using a text generation language called Step~\cite{horswill2022step} and embedding it in a Unity game. StoryAssembler enabled us to split topic threads into storylets and dynamically construct choice sets for these storylets. StoryAssembler models story goals to help with the choice selection process, and we use this to ensure completion of pedagogical goals (see Figure~\ref{fig:storyassembler-example}).

% For example, in version 2.0, once the player learns about “protecting vulnerable persons” as a component of RCR, the system can mix in dialogue options for them to explore that information further.

\section{Method}
To investigate how DENs impact players' engagement and motivation, we conducted a between-subject experimental study comparing 1.0 (static narrative) to 2.0 (dynamic narrative).
%, a newer version of the game that dynamically adapts game content based on players' choices.
Specifically, we aimed to explore how player engagement and ;motivations within the games would differ. This study was approved by the university Institutional Review Board.

\subsection{Procedure}
Participants were required to complete a pre-test questionnaire, during which a unique player ID was assigned to track their gameplay. Following this, they were randomly assigned to one of the two versions of Academical. Total gameplay lasted approximately 20 minutes. Upon completing the assigned version of the game, participants were asked to complete a post-test questionnaire. 

\subsection{Participation}

This study was open to anyone 18 years or older with access to a PC. Participants were recruited through various channels, including Reddit subchannels r/SampleSize and r/TakeMySurvey, as well as the school department’s Slack and Discord channels. Additionally, the study recruitment announcement was distributed to undergraduate summer quarter classes in the school department.
%Consent for participation was addressed and digital signature was required in the beginning of the pre-test questionnaire.
For 1.0, we recruited 49 participants, with 28 completing the study (mean age: 28, SD: 2.85, 16 male, 9 female, 2 non-binary, and 1 preferring not to answer). For 2.0,  we recruited 69 participants, of which 22 completed the study (mean age: 20.9, SD: 3.37, 17 male, 5 female).
%Mean age of participants was 20.52 (SD=3.072), with 33 male, 14 female, 2 non-binary, and 1 answering prefer not to answer. 7 out of 50 participants had RCR training within the past 2 years. 

\subsection{Measurements}

\subsubsection{Questionnaire}

The pre-test questionnaire included questions on basic demographic information, general gameplay experience, and prior RCR training experience. The post-test questionnaire presented two open-ended questions asking participants (1) what they enjoyed the most about their game experience, and (2) what they enjoyed the least. %Specifically, the open-ended questions were: (1) What did you enjoy MOST about your experience with the game? and (2) What did you enjoy LEAST about your experience with the game?

\subsubsection{Qualitative data analysis}

After gathering qualitative data, we familiarized ourselves with it, then generated codes. Two researchers each with a background in Computer Science and Game Design at a technical institution repeatedly reviewed the codes and discussed the reasons for any discrepancies. Through thematic coding \cite{ClarkeBraun2017}, we defined themes that emerged from the coded data. For the coding process, we started with a deductive approach using codes based on the motivational ARC factors. We then conducted a second pass of the data using inductive coding to identify other relevant concepts that emerged from the qualitative data. Ultimately, we identified four additional keywords related to gameplay: `aesthetics,' `narrative,' `simplicity,' and `loop' as shown in Appendix~\ref{code:appendix}. While the definition of `aesthetics' in games research may vary, ranging from sensory content to art forms \cite{bateman2014empirical, niedenthal2009what}, we used `aesthetics' to refer to participants' mentions of visual  elements, such as game graphics. The code `narrative' was used to address storytelling \cite{miller2019digital}, focusing on events \cite{simons2007narrative} and characters \cite{hefner2007identification}. The codes `simplicity' and `loop' mark participants' comments on the simplicity of the game content and a sense of repeated actions respectively. Finally, we did a third pass of all codes in order to generate relevant themes.

\section{Results}

\begin{table*}
  \caption{Code usage in positive and negative comments on Academical 1.0 and 2.0}
  \label{tab:code-usage}
  \Description{Code usage in positive and negative comments on Academical 1.0 and 2.0. In order of Academical 1.0's positive, negative, and Academical 2.0's positive, then negative: Competence: 4 (14.2\%) 5 (17.9\%) 6 (27.2\%) 0 (0\%). Autonomy: 7 (25\%) 4 (14.2\%) 5 (22.7\%) 0 (0\%). Relatedness: 5 (17.9\%) 1 (3.6\%) 6 (27.2\%) 0 (0\%). Aesthetics 10 (35.7\%) 3 (10.7\%) 2 (9.1\%) 1 (4.5\%). Narrative: 9 (32.1\%) 5 (17.9\%) 7 (31.8\%) 4 (18.2\%). Simplicity: 0 (0\%) 7 (2\%) 0 (0\%) 0 (0\%). Loop: 0 (0\%) 0 (0\%) 0 (0\%) 11 (50\%).}
  \centering
  \begin{tabular}{p{0.2\textwidth}p{0.12\textwidth}p{0.1\textwidth}p{0.1\textwidth}p{0.1\textwidth}p{0.1\textwidth}}
    \toprule
   \textbf{Category} & \textbf{Codes} & \multicolumn{2}{c}{\textbf{Academical 1.0}} & \multicolumn{2}{c}{\textbf{Academical 2.0}} \\
                   &                     & \textbf{Positive} & \textbf{Negative} & \textbf{Positive} & \textbf{Negative} \\
    \midrule
    \textbf{Motivational factors} & Competence  & 4 (14.2\%)  & 7 (25\%)  & 8 (36.4\%)  & 7 (31.8\%)  \\
                           & Autonomy    & 9 (32.1\%)    & 4 (14.2\%)  & 6 (27.2\%)  & 2 (9.1\%)  \\
                           & Relatedness & 5 (17.9\%)  & 1 (3.6\%)   & 7 (31.8\%)  & 1 (4.5\%)  \\
    \textbf{Game elements}  & Aesthetics  & 10 (35.7\%) & 3 (10.7\%)  & 2 (9.1\%)   & 1 (4.5\%) \\
                           & Narrative   & 9 (32.1\%)  & 5 (17.9\%)  & 7 (31.8\%)  & 4 (18.2\%) \\
    \textbf{Perception}     & Simplicity  & 0 (0\%)     & 7 (25\%)    & 0 (0\%)     & 0 (0\%) \\
                           & Loop        & 0 (0\%)     & 0 (0\%)     & 0 (0\%)     & 11 (50\%) \\
    \bottomrule
  \end{tabular}
\end{table*}

\subsection{Coding}

The frequency of each code mentioned in the participants' comments is shown in Table~\ref{tab:code-usage}. While the same codes were used to mark comments for both 1.0 and 2.0, covering both positive and negative aspects, ‘loop’ appeared only in 2.0 as a key negative element, whereas ‘simplicity’ appeared only in 1.0 as a key negative element. The frequent mention of simplicity in 1.0 seems to be due to its quick transition to failure or success, whereas 2.0 may have been perceived as having richer content due to its dynamic nature.
%P16 in the 1.0 condition specifically addressed \textit{“the lack of content”} when asked what they enjoyed the least.
Looping was an issue raised by many participants in version 2.0 as the dynamic implementation required players to revisit and navigate through sequences again if a certain learning goal was not completed, leading to feelings of "looping".

Turning to the motivational factor codes, both 1.0 and 2.0 included mentions of all three codes---competence, autonomy, and relatedness. Interestingly, these codes were mentioned primarily in a positive context for 2.0, while for 1.0, they were addressed in both positive and negative contexts. %Even ‘competence’ was mentioned more frequently and positively with version 2.0. This could be due to the limited scope of the competence-related questions in the PENS questionnaire, or perhaps the perceived loop issue in version 2.0 overshadowed other perspectives.

For ‘narrative,’ the percentage of the code being applied was roughly the same for both versions across positive and negative perspectives. There was a more frequent mention of ‘aesthetic’ as a positive aspect in 1.0, which may be due to the perceived simplicity of its user interface. Meanwhile, the negative comments on ‘aesthetics’ for both versions were more about personal preference for graphic styles.

\subsection{Thematic Analysis}
Our thematic analysis identified 3 key themes that arose from participant responses.

\subsubsection{Theme 1 - Dynamism and Learning Goals}
Much of learning is driven by exploration and experimentation. Procedural narrative systems give players the flexibility to explore while providing the game creator a means of control over the narrative. Notably, the dynamic nature of dialogue sequencing in 2.0 seemed capable of giving players the feeling of a possibility space that they could explore. This is evidenced by participant comments such as, \textit{“This game has a lot of options, and each option has a different branch, and each branch has a different dialogue content, which makes me feel that this game can be fully carried out.”} (P2 about the most enjoyable point of 2.0).

However, the balance between learning goals and dynamic narrative is a difficult one to strike. Dynamic narratives provide far more content to traverse and explore, theoretically enhancing the feelings of autonomy and competence experienced during play. However, the requirement to simultaneously meet learning goals and limit narrative progression based upon completion of these goals may have limited the strengths that dynamism brings to narrative. The result of limiting narrative progression based on learning progress was a lack of clear narrative repercussions for the player. Specifically, 1.0 featured short scenarios that ended quickly in success and failure cases, while 2.0 focused on a longer scenario with sequencing of pedagogical goals that could not be clearly failed. Instead, players would be required to ``loop back'' if a given learning outcome was not completed. This led to a sense of confusion and repetitiveness that negatively impacted players' feelings of competence. As one participant put it, \textit{"some of the content chose different options and circled back to key situations that made me feel like I was treading water instead of moving forward, and there were always places that felt difficult"} (P42, 2.0). Furthermore, players had no clear way of understanding their progress through learning objectives or if they were close to resolving the narrative. As noted by P48, \textit{"...Sometimes it felt like I was going in a loop and I was trying to figure out how to progress forward in the narrative which breaks the immersive experience."} This is less problematic in the short, punchy narratives of 1.0, but considering the length of 2.0, this contributed to a feeling of sluggish progression.

\subsubsection{Theme 2 - Responsive Graphical Feedback}
As players made dialogue choices, they received different visual reactions from the characters (e.g., changes in facial expression). These aesthetics in 2.0 seemed to have a notable connection with competence and relatedness. For instance, the aesthetics helped players place themselves in the characters' shoes and perceive their interaction with characters as their own, e.g., \textit{"Illustrations were great too; the facial expression hurt"} (P50, 2.0). Additionally, aesthetics supported competence as well since players perceived changes in character illustrations as direct feedback for their decisions, e.g., \textit{"I enjoyed seeing the reactions of the characters face everytime I choose a option"} (P47, 2.0).

\subsubsection{Theme 3 - Consequential choices}
Within 2.0, we did not provide players with choices that could result in meaningful future consequences. We focused more on providing choices that allowed the player to explore educational topics in varying orders and depths without consequence. However, without a way to gauge the long-term effects of their choices, players had difficulty determining when and where they made progress. For instance, P29 commented that the least enjoyable aspect of 2.0 is the \textit{“Lack of clarity on what is considered progress. Cyclical questions”}. In traditional static narratives, in order to experience autonomy, players must be presented with consequential choices, they must have a basis for hypothesizing why consequences will occur, and they must receive feedback that allows them to adjust their future hypotheses \cite{wardripfruin2009agency}. Ultimately, it appears that the increase in dynamism which DENs provide was unable to compensate for the lack of consequential choices within the narrative.

\section{Design Implications}

\subsection{Responsive content}
One unexpected insight from our study was the importance of paring the wide range of content that DENs afford with responsive feedback that gives players a clearer sense of progression. Specifically, both responsive graphical feedback and meaningful choices (or lack thereof) had direct impact on players' feelings of autonomy, relatedness, and competency during the game. Therefore, for the design of DENs, it is crucial to provide responsive graphical feedback and frequent consequential choices within the narrative in order to clarify progression and enhance player motivation.

\subsection{Avoid Looping and Repeating Dialogue Hubs}
%We hypothesize that our attempts to limit progression based upon pedagogical goals ultimately reduced opportunities to utilize more systems-driven approaches to the narrative progression. Despite  substantially more content and choices available within Academical 2.0, players' perceptions of autonomy and competence was limited due to this unseen pedagogical barrier to progression. 
A common point of needs frustration in 2.0 was the repeating of certain specific utterances when conversation or pedagogical requirements were not properly fulfilled, leading to a feeling of ``looping.'' P43 said \textit{"I tested a lot of the options and they either repeat or lead back to something I have already done which is a little annoying."} This pattern of narrative is referred to as the ``Spoke and Hub'' pattern, which has a central node that is returned to after different branches of content have been visited and completed \cite{millard2023strange}. Spoke and Hub approaches are common in both narrative and spatial game design, featuring a central area with sub-areas that must be explored before the central area can be ``solved'' \cite{wei2010time}. While it is possible this pattern can produce good results within dynamic educational interactive narratives, our findings indicate that Spoke and Hub approaches impacted players' feelings of competence and autonomy by forcing them through the same dialogue points repeatedly.
%This outcome may have been mitigated if multiple alternate Hub nodes were provided, effectively hiding the repeating of branching points. However,
The removal of clear outcomes to choices suggests that this pattern should mostly be avoided (or implemented with care and cleverness) in DENs, where learning is linked to a constant sense of progression and repercussions.
%Despite their widespread use, Spoke and Hub patterns can have substantial drawbacks to gameplay experience. The \textit{Game Developer} guide to branching conversations states that players may "feel obligated to punch every button and get each available tidbit of data," which would ultimately lose natural flow.

% ``At their best, hubs make Players feel clever and in charge, picking and choosing how to approach their subjects in interrogation. At their worst, hubs turn NPCs into information vending machines where the Player feels obligated to punch every button and get each available tidbit of data. Such conversations lose any sort of naturalistic flow; pacing ceases to exist and...actors veer wildly from one subject to the next.'' \cite{Freed_2014} 

\subsection{Role-Playing Opportunities}

A key component during the design of 2.0 was the interweaving of non-pedagogical content that involved humor, drama, or negative choices. Writers were encouraged to include this type of content along with learning content to offer pacing breaks and to increase the realism of characters that might naturally joke or feel insecure in response to difficult topics. Role-playing, or the ability to occupy a character with ease, was a goal of the project's dynamic narrative approach. Based on the qualitative data, this seems to have been successful for players. It increased the perceived relatedness of players, as they could more easily put themselves in the characters' shoes: \textit{''I liked the narrative part a lot, because I felt like I was learning about the IRB ... I felt like I was having a full conversation with Ned. It was immersive as well, for a minute I felt the same stress as Brad did while trying to talk to Ned"} (P47, 2.0). Additionally, while narrative games mostly progress through making decisions among plausible choices provided within the game, players often choose options not only perceived as `correct answers' but also based on their personal preferences. As P33 put it, the aspect of version 2.0 that they most enjoyed was, \textit{``Clicking some of the silly options''} which 2.0 had in abundance due the greater amount of content and choices provided at each point in the game. In the serious games literature, this interplay between gameplay and narrative entertainment with learning outcomes is referred to as ``balance'' \cite{frank2007balancing}. This body of literature emphasizes that inserting types of entertainment and role-play are important to ensuring learning outcomes as they enable player pacing between learning and entertainment tasks. For example, \citet{hall2014instructional} indicates that balancing core gameplay with pedagogy helps avoid making ``chocolate-dipped broccoli''---where learning outcomes are lightly covered with a game facade to force positive outcomes, resulting in an unsatisfying media product. This lesson is echoed by \citet{johansson2014design}, who highlights that finding this entertainment-learning balance maximizes learning and should be a focus of the design process of serious games. As such, DENs should make sure to incorporate entertaining gameplay options for players to explore while interweaving educational content to maximize motivation.

\subsection{Limitations and Future Work}

The attempt to compare a single scenario in 2.0 to an existing scenario in 1.0 hampered the potential dynamism and perspective taking that could have increased autonomy, relatedness, and competence. While we scoped the game design and study this way to account for increased writing complexity and to create a comparison point, it may have ultimately limited the full breadth of narrative and educational affordances a DEN approach provides.
%it was not immediately apparent how shorter, more diverse scenarios could provide a sense of progress and feedback for the players.
Additionally, study retention was challenging as the full procedure was complicated for participants to follow and required multiple steps (pre-survey, game download and play, post-survey, all while tracking an identifier throughout) to complete. While we did have attrition on the study, it appears to be due to the study's complexity rather than the gameplay experience itself. For example, in the Academical 2.0 condition, only 25\% of participants did not finish the story completely, and even they spent significant time playing the game. The shortest playtime of those who did not complete the game was 11 minutes with 51 dialogue choices made. This indicates that retention issues were related to the procedural complexity of the study rather than dissatisfaction with the gameplay. Another limitation of the study is the lack of quantitative measurements to support the qualitative insights provided here - future work will aim to build on these findings by incorporating quantitative methods to help better inform qualitative results.

Future work will also involve the creation of a new DEN system that will address the design weaknesses identified above. Specifically, the updated system will feature shorter scenes that have clearly defined character and pedagogical outcomes that can be observed by the player. These will still be sequenced to maintain the strengths of interactive digital narrative practices. We will also continue to incorporate a mix of non-learning and learning content, providing goals in both categories to maintain the strengths of player role-playing balance that proved effective in 2.0.
%Similarly, the gameplay length for Academical 2.0 was longer than 1.0, as its dynamic nature prevented quick failure cases. This may have led to a higher dropout rate among participants assigned to Academical 2.0. The inherent differences between static and dynamic narratives may have also inevitably affected the overall gameplay experience. 

 % Since this study required participants to complete a pre-test questionnaire, play the assigned version of the game for approximately 20 minutes, and then complete a post-test questionnaire, retention issues may have occurred. This could have contributed to the difficulty in recruiting larger sample as well. Academical 2.0 required downloading the game file, rather than web-based access, due to its dynamic design implementation. 
 
 % A final limitation is the retention issues of the study. The study process was lengthy and featured multiple steps --- players would complete an initial survey, be linked to a website to download or play a web version, then be linked to a final survey. This required the tracking of player IDs throughout the process, a task that proved overwhelmingly difficult for participants. 

% \subsection{Future Work}

% Participants often listed the looping narrative structure as something that contributed negatively to their gameplay experience. Additionally, some participants also stated they felt lost since the second version of the game did not provide them with any immediate goal(s).

\section{Conclusion}

This paper investigated the impacts on player engagement and motivation for static and dynamic educational interactive narratives about RCR. We found that 1) there are notable challenges balancing the dynamism of DENs which can impact feelings of autonomy and competence, 2) responsive graphical feedback is valuable for fostering feelings of relatedness and competence, and 3) consequential choices are crucial for DENs despite the much broader range of content and choices overall. Future DEN designs should focus on creating responsive content and consequential choices for players, avoiding looping and repeating dialogue hubs, and providing opportunities for silly options among the serious.

\begin{acks}
This material is based upon work supported by the National Science Foundation under Grant No. 2202521. Any opinions, findings, and conclusions or recommendations expressed in this material are those of the author(s) and do not necessarily reflect the views of the National Science Foundation. This work was made possible by the contributors of the project: Celine Lafosse, Sophia Nguyen, Rachel Kadem, and Hebah Haque. %%TODO - Writers, Developers, Artists
\end{acks}

\bibliographystyle{ACM-Reference-Format}
\bibliography{sample-base.bib} 


\begin{thebibliography}{61}

%%% the Web address.

\ifx \showCODEN    \undefined \def \showCODEN     #1{\unskip}     \fi
\ifx \showISBNx    \undefined \def \showISBNx     #1{\unskip}     \fi
\ifx \showISBNxiii \undefined \def \showISBNxiii  #1{\unskip}     \fi
\ifx \showISSN     \undefined \def \showISSN      #1{\unskip}     \fi
\ifx \showLCCN     \undefined \def \showLCCN      #1{\unskip}     \fi
\ifx \shownote     \undefined \def \shownote      #1{#1}          \fi
\ifx \showarticletitle \undefined \def \showarticletitle #1{#1}   \fi
\ifx \showURL      \undefined \def \showURL       {\relax}        \fi
% invisible to TeX
\providecommand\bibfield[2]{#2}
\providecommand\bibinfo[2]{#2}
\providecommand\natexlab[1]{#1}
\providecommand\showeprint[2][]{arXiv:#2}

\bibitem[Agusalim(2012)]%
        {Agusalim2012}
\bibfield{author}{\bibinfo{person}{Imam~Dui Agusalim}.} \bibinfo{year}{2012}\natexlab{}.
\newblock \showarticletitle{Developing Interactive E-Learning Module of English Teaching to Support the Distance Education Program at EEPIS}.
\newblock \bibinfo{journal}{\emph{IOSR Journal of Humanities and Social Science}} \bibinfo{volume}{5}, \bibinfo{number}{1} (\bibinfo{year}{2012}), \bibinfo{pages}{28--32}.
\newblock
\href{https://doi.org/10.9790/0837-0512832}{doi:\nolinkurl{10.9790/0837-0512832}}


\bibitem[Amalo et~al\mbox{.}(2017)]%
        {Amalo2017}
\bibfield{author}{\bibinfo{person}{E.~A. Amalo}, \bibinfo{person}{I.~D. Agusalim}, {and} \bibinfo{person}{C.~D. Murdaningtyas}.} \bibinfo{year}{2017}\natexlab{}.
\newblock \showarticletitle{Developing visual novel game with speech-recognition interactivity to enhance students' mastery on English expressions}.
\newblock \bibinfo{journal}{\emph{Jurnal Sosial Humaniora}} \bibinfo{volume}{10}, \bibinfo{number}{2} (\bibinfo{year}{2017}), \bibinfo{pages}{129}.
\newblock
\href{https://doi.org/10.12962/j24433527.v10i2.2865}{doi:\nolinkurl{10.12962/j24433527.v10i2.2865}}


\bibitem[Andrew et~al\mbox{.}(2019)]%
        {Andrew2019}
\bibfield{author}{\bibinfo{person}{J. Andrew}, \bibinfo{person}{S. Henry}, \bibinfo{person}{A.~N. Yudhisthira}, \bibinfo{person}{Y. Arifin}, {and} \bibinfo{person}{S.~D. Permai}.} \bibinfo{year}{2019}\natexlab{}.
\newblock \showarticletitle{Analyzing the factors that influence learning experience through game-based learning using a visual novel game for learning pancasila}.
\newblock \bibinfo{journal}{\emph{Procedia Computer Science}}  \bibinfo{volume}{157} (\bibinfo{year}{2019}), \bibinfo{pages}{353--359}.
\newblock
\href{https://doi.org/10.1016/j.procs.2019.08.177}{doi:\nolinkurl{10.1016/j.procs.2019.08.177}}


\bibitem[Bateman(2014)]%
        {bateman2014empirical}
\bibfield{author}{\bibinfo{person}{Chris Bateman}.} \bibinfo{year}{2014}\natexlab{}.
\newblock \showarticletitle{Empirical Game Aesthetics}.
\newblock In \bibinfo{booktitle}{\emph{Handbook of Digital Games}}, \bibfield{editor}{\bibinfo{person}{Marios~C. Angelides} {and} \bibinfo{person}{Harry Agius}} (Eds.). \bibinfo{publisher}{John Wiley \& Sons, Inc.}, \bibinfo{pages}{Chapter 15}.
\newblock


\bibitem[Bond et~al\mbox{.}(2020)]%
        {bond2020mapping}
\bibfield{author}{\bibinfo{person}{Melissa Bond}, \bibinfo{person}{Katja Buntins}, \bibinfo{person}{Svenja Bedenlier}, \bibinfo{person}{Olaf Zawacki-Richter}, {and} \bibinfo{person}{Michael Kerres}.} \bibinfo{year}{2020}\natexlab{}.
\newblock \showarticletitle{Mapping research in student engagement and educational technology in higher education: A systematic evidence map}.
\newblock \bibinfo{journal}{\emph{International journal of educational technology in higher education}}  \bibinfo{volume}{17} (\bibinfo{year}{2020}), \bibinfo{pages}{1--30}.
\newblock


\bibitem[Bouville(2008)]%
        {bouville2008}
\bibfield{author}{\bibinfo{person}{Mathieu Bouville}.} \bibinfo{year}{2008}\natexlab{}.
\newblock \showarticletitle{On using ethical theories to teach engineering ethics}.
\newblock \bibinfo{journal}{\emph{Science and Engineering Ethics}} \bibinfo{volume}{14}, \bibinfo{number}{1} (\bibinfo{year}{2008}), \bibinfo{pages}{111--120}.
\newblock


\bibitem[Brummel et~al\mbox{.}(2010a)]%
        {brummel2010}
\bibfield{author}{\bibinfo{person}{Bradley~J Brummel}, \bibinfo{person}{CK Gunsalus}, \bibinfo{person}{Kerri~L Anderson}, {and} \bibinfo{person}{Michael~C Loui}.} \bibinfo{year}{2010}\natexlab{a}.
\newblock \showarticletitle{Development of role-play scenarios for teaching responsible conduct of research}.
\newblock \bibinfo{journal}{\emph{Science and Engineering Ethics}} \bibinfo{volume}{16}, \bibinfo{number}{3} (\bibinfo{year}{2010}), \bibinfo{pages}{573--589}.
\newblock


\bibitem[Brummel et~al\mbox{.}(2010b)]%
        {brummel2010development}
\bibfield{author}{\bibinfo{person}{Bradley~J Brummel}, \bibinfo{person}{CK Gunsalus}, \bibinfo{person}{Kerri~L Anderson}, {and} \bibinfo{person}{Michael~C Loui}.} \bibinfo{year}{2010}\natexlab{b}.
\newblock \showarticletitle{Development of role-play scenarios for teaching responsible conduct of research}.
\newblock \bibinfo{journal}{\emph{Science and engineering ethics}}  \bibinfo{volume}{16} (\bibinfo{year}{2010}), \bibinfo{pages}{573--589}.
\newblock


\bibitem[Camingue et~al\mbox{.}(2021)]%
        {Camingue2021}
\bibfield{author}{\bibinfo{person}{Janelynn Camingue}, \bibinfo{person}{Elin Carstensdottir}, {and} \bibinfo{person}{Edward~F. Melcer}.} \bibinfo{year}{2021}\natexlab{}.
\newblock \showarticletitle{What is a visual novel?}
\newblock \bibinfo{journal}{\emph{Proceedings of the ACM on Human-Computer Interaction}} \bibinfo{volume}{5}, \bibinfo{number}{CHI PLAY} (\bibinfo{year}{2021}), \bibinfo{pages}{1--18}.
\newblock
\href{https://doi.org/10.1145/3474712}{doi:\nolinkurl{10.1145/3474712}}


\bibitem[Camingue et~al\mbox{.}(2020)]%
        {Camingue2020}
\bibfield{author}{\bibinfo{person}{Janelynn Camingue}, \bibinfo{person}{Edward~F. Melcer}, {and} \bibinfo{person}{Elin Carstensdottir}.} \bibinfo{year}{2020}\natexlab{}.
\newblock \showarticletitle{A (Visual) Novel Route to Learning: A Taxonomy of Teaching Strategies in Visual Novels}. In \bibinfo{booktitle}{\emph{International Conference on the Foundations of Digital Games (FDG '20)}} (Bugibba, Malta). \bibinfo{publisher}{Association for Computing Machinery}, \bibinfo{pages}{Article 77, 13 pages}.
\newblock
\href{https://doi.org/10.1145/3402942.3403004}{doi:\nolinkurl{10.1145/3402942.3403004}}


\bibitem[Clarke and Braun(2017)]%
        {ClarkeBraun2017}
\bibfield{author}{\bibinfo{person}{V. Clarke} {and} \bibinfo{person}{V. Braun}.} \bibinfo{year}{2017}\natexlab{}.
\newblock \showarticletitle{Thematic analysis}.
\newblock \bibinfo{journal}{\emph{The Journal of Positive Psychology}} \bibinfo{volume}{12}, \bibinfo{number}{3} (\bibinfo{year}{2017}), \bibinfo{pages}{297--298}.
\newblock
\href{https://doi.org/10.1080/17439760.2016.1262613}{doi:\nolinkurl{10.1080/17439760.2016.1262613}}


\bibitem[Faizal(2016)]%
        {Faizal2016}
\bibfield{author}{\bibinfo{person}{M.~Aliv Faizal}.} \bibinfo{year}{2016}\natexlab{}.
\newblock \showarticletitle{The effects of conversation-gambits visual-novel game on students' English achievement and motivation}. In \bibinfo{booktitle}{\emph{2016 International Electronics Symposium (IES)}}. \bibinfo{publisher}{IEEE}, \bibinfo{pages}{481--486}.
\newblock
\href{https://doi.org/10.1109/elecsym.2016.7861054}{doi:\nolinkurl{10.1109/elecsym.2016.7861054}}


\bibitem[Frank(2007)]%
        {frank2007balancing}
\bibfield{author}{\bibinfo{person}{Anders Frank}.} \bibinfo{year}{2007}\natexlab{}.
\newblock \showarticletitle{Balancing three different foci in the design of serious games: Engagement, training objective and context}. In \bibinfo{booktitle}{\emph{Proceedings of DiGRA 2007 Conference: Situated Play}}.
\newblock


\bibitem[Friedhoff(2013)]%
        {friedhoff2013}
\bibfield{author}{\bibinfo{person}{Jane Friedhoff}.} \bibinfo{year}{2013}\natexlab{}.
\newblock \showarticletitle{Untangling Twine: A Platform Study}. In \bibinfo{booktitle}{\emph{Proc. DiGRA}}.
\newblock


\bibitem[Gabriel et~al\mbox{.}(2018)]%
        {Gabriel2018}
\bibfield{author}{\bibinfo{person}{Pedro Gabriel}, \bibinfo{person}{Tsukasa Hirashima}, {and} \bibinfo{person}{Hayashi Yusuke}.} \bibinfo{year}{2018}\natexlab{}.
\newblock \showarticletitle{A Serious Game for Improving Inferencing in the Presence of Foreign Language Unknown Words}.
\newblock \bibinfo{journal}{\emph{International Journal of Advanced Computer Science and Applications}} \bibinfo{volume}{9}, \bibinfo{number}{2} (\bibinfo{year}{2018}), \bibinfo{pages}{341}.
\newblock
\href{https://doi.org/10.14569/ijacsa.2018.090202}{doi:\nolinkurl{10.14569/ijacsa.2018.090202}}


\bibitem[Garbe et~al\mbox{.}(2019)]%
        {garbe2019storyassembler}
\bibfield{author}{\bibinfo{person}{Jacob Garbe}, \bibinfo{person}{Max Kreminski}, \bibinfo{person}{Ben Samuel}, \bibinfo{person}{Noah Wardrip-Fruin}, {and} \bibinfo{person}{Michael Mateas}.} \bibinfo{year}{2019}\natexlab{}.
\newblock \showarticletitle{StoryAssembler: an engine for generating dynamic choice-driven narratives}. In \bibinfo{booktitle}{\emph{Proceedings of the 14th International Conference on the Foundations of Digital Games}}. \bibinfo{pages}{1--10}.
\newblock


\bibitem[G{\"o}bel et~al\mbox{.}(2009)]%
        {gobel200980days}
\bibfield{author}{\bibinfo{person}{Stefan G{\"o}bel}, \bibinfo{person}{Florian Mehm}, \bibinfo{person}{Sabrina Radke}, {and} \bibinfo{person}{Ralf Steinmetz}.} \bibinfo{year}{2009}\natexlab{}.
\newblock \showarticletitle{80days: Adaptive digital storytelling for digital educational games}. In \bibinfo{booktitle}{\emph{Proceedings of the 2nd international workshop on Story-Telling and Educational Games (STEG’09)}}, Vol.~\bibinfo{volume}{498}.
\newblock


\bibitem[Grasse et~al\mbox{.}(2022)]%
        {grasse2022using}
\bibfield{author}{\bibinfo{person}{Katelyn~M Grasse}, \bibinfo{person}{Max Kreminski}, \bibinfo{person}{Noah Wardrip-Fruin}, \bibinfo{person}{Michael Mateas}, {and} \bibinfo{person}{Edward~F Melcer}.} \bibinfo{year}{2022}\natexlab{}.
\newblock \showarticletitle{Using self-determination theory to explore enjoyment of educational interactive narrative games: a case study of academical}.
\newblock \bibinfo{journal}{\emph{Frontiers in Virtual Reality}}  \bibinfo{volume}{3} (\bibinfo{year}{2022}), \bibinfo{pages}{847120}.
\newblock


\bibitem[Grasse et~al\mbox{.}(2021b)]%
        {grasse2021academical}
\bibfield{author}{\bibinfo{person}{Katelyn~M Grasse}, \bibinfo{person}{Edward~F Melcer}, \bibinfo{person}{Max Kreminski}, \bibinfo{person}{Nick Junius}, \bibinfo{person}{James Ryan}, {and} \bibinfo{person}{Noah Wardrip-Fruin}.} \bibinfo{year}{2021}\natexlab{b}.
\newblock \showarticletitle{Academical: a choice-based interactive storytelling game for enhancing moral reasoning, knowledge, and attitudes in responsible conduct of research}.
\newblock In \bibinfo{booktitle}{\emph{Games and Narrative: Theory and Practice}}. \bibinfo{publisher}{Springer}, \bibinfo{pages}{173--189}.
\newblock


\bibitem[Grasse et~al\mbox{.}(2021a)]%
        {grasse2021improving}
\bibfield{author}{\bibinfo{person}{Katelyn~M Grasse}, \bibinfo{person}{Edward~F Melcer}, \bibinfo{person}{Max Kreminski}, \bibinfo{person}{Nick Junius}, {and} \bibinfo{person}{Noah Wardrip-Fruin}.} \bibinfo{year}{2021}\natexlab{a}.
\newblock \showarticletitle{Improving undergraduate attitudes towards responsible conduct of research through an interactive storytelling game}. In \bibinfo{booktitle}{\emph{Extended abstracts of the 2021 CHI conference on human factors in computing systems}}. \bibinfo{pages}{1--8}.
\newblock


\bibitem[Guay et~al\mbox{.}(2008)]%
        {guay2008optimal}
\bibfield{author}{\bibinfo{person}{Fr{\'e}d{\'e}ric Guay}, \bibinfo{person}{Catherine~F Ratelle}, {and} \bibinfo{person}{Julien Chanal}.} \bibinfo{year}{2008}\natexlab{}.
\newblock \showarticletitle{Optimal learning in optimal contexts: The role of self-determination in education.}
\newblock \bibinfo{journal}{\emph{Canadian psychology/Psychologie canadienne}} \bibinfo{volume}{49}, \bibinfo{number}{3} (\bibinfo{year}{2008}), \bibinfo{pages}{233}.
\newblock


\bibitem[Hall et~al\mbox{.}(2014)]%
        {hall2014instructional}
\bibfield{author}{\bibinfo{person}{Joshua~V Hall}, \bibinfo{person}{Peta~A Wyeth}, {and} \bibinfo{person}{Daniel Johnson}.} \bibinfo{year}{2014}\natexlab{}.
\newblock \showarticletitle{Instructional objectives to core-gameplay: a serious game design technique}. In \bibinfo{booktitle}{\emph{Proceedings of the first ACM SIGCHI annual symposium on Computer-human interaction in play}}. \bibinfo{pages}{121--130}.
\newblock


\bibitem[Harrell and Zhu(2009)]%
        {harrell2009agency}
\bibfield{author}{\bibinfo{person}{D~Fox Harrell} {and} \bibinfo{person}{Jichen Zhu}.} \bibinfo{year}{2009}\natexlab{}.
\newblock \showarticletitle{Agency Play: Dimensions of Agency for Interactive Narrative Design.}. In \bibinfo{booktitle}{\emph{AAAI spring symposium: Intelligent narrative technologies II}}. \bibinfo{pages}{44--52}.
\newblock


\bibitem[Hefner et~al\mbox{.}(2007)]%
        {hefner2007identification}
\bibfield{author}{\bibinfo{person}{Doroth{\'e}e Hefner}, \bibinfo{person}{Christoph Klimmt}, {and} \bibinfo{person}{Peter Vorderer}.} \bibinfo{year}{2007}\natexlab{}.
\newblock \showarticletitle{Identification with the Player Character as Determinant of Video Game Enjoyment}.
\newblock In \bibinfo{booktitle}{\emph{Entertainment Computing – ICEC 2007}}, \bibfield{editor}{\bibinfo{person}{Lizhu Ma}, \bibinfo{person}{Matthias Rauterberg}, {and} \bibinfo{person}{Ryohei Nakatsu}} (Eds.). \bibinfo{series}{Lecture Notes in Computer Science}, Vol.~\bibinfo{volume}{4740}. \bibinfo{publisher}{Springer, Berlin, Heidelberg}, \bibinfo{pages}{39--48}.
\newblock
\href{https://doi.org/10.1007/978-3-540-74873-1_6}{doi:\nolinkurl{10.1007/978-3-540-74873-1_6}}


\bibitem[Horswill(2022)]%
        {horswill2022step}
\bibfield{author}{\bibinfo{person}{Ian Horswill}.} \bibinfo{year}{2022}\natexlab{}.
\newblock \showarticletitle{Step: a highly expressive text generation language}. In \bibinfo{booktitle}{\emph{Proceedings of the AAAI Conference on Artificial Intelligence and Interactive Digital Entertainment}}, Vol.~\bibinfo{volume}{18}. \bibinfo{pages}{240--249}.
\newblock


\bibitem[Hsu et~al\mbox{.}(2019)]%
        {hsu2019reexamining}
\bibfield{author}{\bibinfo{person}{Hui-Ching~Kayla Hsu}, \bibinfo{person}{Cong~Vivi Wang}, {and} \bibinfo{person}{Chantal Levesque-Bristol}.} \bibinfo{year}{2019}\natexlab{}.
\newblock \showarticletitle{Reexamining the impact of self-determination theory on learning outcomes in the online learning environment}.
\newblock \bibinfo{journal}{\emph{Education and information technologies}} \bibinfo{volume}{24}, \bibinfo{number}{3} (\bibinfo{year}{2019}), \bibinfo{pages}{2159--2174}.
\newblock


\bibitem[Huynh et~al\mbox{.}(2017)]%
        {Huynh2017}
\bibfield{author}{\bibinfo{person}{Duy Huynh}, \bibinfo{person}{Phuc Luong}, \bibinfo{person}{Hiroyuki Iida}, {and} \bibinfo{person}{Razvan Beuran}.} \bibinfo{year}{2017}\natexlab{}.
\newblock \showarticletitle{Design and Evaluation of a Cybersecurity Awareness Training Game}. In \bibinfo{booktitle}{\emph{Entertainment Computing -- ICEC 2017}}. \bibinfo{publisher}{Springer International Publishing, Cham}, \bibinfo{pages}{183--188}.
\newblock


\bibitem[Iten et~al\mbox{.}(2018)]%
        {iten2018choosing}
\bibfield{author}{\bibinfo{person}{Glena~H Iten}, \bibinfo{person}{Sharon~T Steinemann}, {and} \bibinfo{person}{Klaus Opwis}.} \bibinfo{year}{2018}\natexlab{}.
\newblock \showarticletitle{Choosing to help monsters: A mixed-method examination of meaningful choices in narrative-rich games and interactive narratives}. In \bibinfo{booktitle}{\emph{Proceedings of the 2018 CHI conference on human factors in computing systems}}. \bibinfo{pages}{1--13}.
\newblock


\bibitem[Johansson et~al\mbox{.}(2014)]%
        {johansson2014design}
\bibfield{author}{\bibinfo{person}{Magnus Johansson}, \bibinfo{person}{Harko Verhagen}, \bibinfo{person}{Anna {\AA}kerfeldt}, {and} \bibinfo{person}{Staffan Selander}.} \bibinfo{year}{2014}\natexlab{}.
\newblock \showarticletitle{How to design for meaningful learning--finding the balance between learning and game components}. In \bibinfo{booktitle}{\emph{Proceedings of the 8th European conference on games based learning}}. \bibinfo{pages}{216--222}.
\newblock


\bibitem[Johnson-Bey et~al\mbox{.}(2024)]%
        {johnson2024academical}
\bibfield{author}{\bibinfo{person}{Shi Johnson-Bey}, \bibinfo{person}{Samuel Shields}, \bibinfo{person}{Noah Wardrip-Fruin}, {and} \bibinfo{person}{Edward Melcer}.} \bibinfo{year}{2024}\natexlab{}.
\newblock \showarticletitle{Academical: A Dynamic Interactive Narrative Game for Responsible Conduct of Research Training}. In \bibinfo{booktitle}{\emph{2024 IEEE Conference on Games (CoG)}}. IEEE, \bibinfo{pages}{1--2}.
\newblock


\bibitem[Kalichman(2013a)]%
        {kalichman2013}
\bibfield{author}{\bibinfo{person}{Michael Kalichman}.} \bibinfo{year}{2013}\natexlab{a}.
\newblock \showarticletitle{A brief history of RCR education}.
\newblock \bibinfo{journal}{\emph{Accountability in Research}} \bibinfo{volume}{20}, \bibinfo{number}{5-6} (\bibinfo{year}{2013}), \bibinfo{pages}{380--394}.
\newblock


\bibitem[Kalichman(2013b)]%
        {kalichman2013b}
\bibfield{author}{\bibinfo{person}{Michael Kalichman}.} \bibinfo{year}{2013}\natexlab{b}.
\newblock \showarticletitle{A brief history of RCR education}.
\newblock \bibinfo{journal}{\emph{Accountability in Research}} \bibinfo{volume}{20}, \bibinfo{number}{5-6} (\bibinfo{year}{2013}), \bibinfo{pages}{380--394}.
\newblock


\bibitem[Kalichman(2014)]%
        {kalichman2014}
\bibfield{author}{\bibinfo{person}{Michael Kalichman}.} \bibinfo{year}{2014}\natexlab{}.
\newblock \showarticletitle{Rescuing responsible conduct of research (RCR) education}.
\newblock \bibinfo{journal}{\emph{Accountability in Research}} \bibinfo{volume}{21}, \bibinfo{number}{1} (\bibinfo{year}{2014}), \bibinfo{pages}{68--83}.
\newblock


\bibitem[Koidl et~al\mbox{.}(2010)]%
        {koidl2010dynamically}
\bibfield{author}{\bibinfo{person}{Kevin Koidl}, \bibinfo{person}{Florian Mehm}, \bibinfo{person}{Cormac Hampson}, \bibinfo{person}{Owen Conlan}, {and} \bibinfo{person}{Stefan G{\"o}bel}.} \bibinfo{year}{2010}\natexlab{}.
\newblock \showarticletitle{Dynamically adjusting digital educational games towards learning objectives}. In \bibinfo{booktitle}{\emph{Proceedings of the Third European Conference on Games Based Learning (ECGBL 2010)}}. \bibinfo{pages}{177--184}.
\newblock


\bibitem[Korhonen and Halonen(2017)]%
        {Korhonen2017}
\bibfield{author}{\bibinfo{person}{Tanja Korhonen} {and} \bibinfo{person}{Raija Halonen}.} \bibinfo{year}{2017}\natexlab{}.
\newblock \showarticletitle{On the Development of Serious Games in the Health Sector - A Case Study of a Serious Game Tool to Improve Life Management Skills in the Young}. In \bibinfo{booktitle}{\emph{Proceedings of the 19th International Conference on Enterprise Information Systems}}.
\newblock
\href{https://doi.org/10.5220/0006331001350142}{doi:\nolinkurl{10.5220/0006331001350142}}


\bibitem[Kreminski and Wardrip-Fruin(2018)]%
        {kreminski2018storylets}
\bibfield{author}{\bibinfo{person}{Max Kreminski} {and} \bibinfo{person}{Noah Wardrip-Fruin}.} \bibinfo{year}{2018}\natexlab{}.
\newblock \showarticletitle{Sketching a Map of the Storylets Design Space}. In \bibinfo{booktitle}{\emph{Interactive Storytelling}}, \bibfield{editor}{\bibinfo{person}{Rebecca Rouse}, \bibinfo{person}{Hartmut Koenitz}, {and} \bibinfo{person}{Mads Haahr}} (Eds.). \bibinfo{publisher}{Springer International Publishing}, \bibinfo{address}{Cham}, \bibinfo{pages}{160--164}.
\newblock
\showISBNx{978-3-030-04028-4}


\bibitem[Lai and Chen(2021)]%
        {Lai2021}
\bibfield{author}{\bibinfo{person}{K.-W.~K. Lai} {and} \bibinfo{person}{H.-J.~H. Chen}.} \bibinfo{year}{2021}\natexlab{}.
\newblock \showarticletitle{A comparative study on the effects of a VR and PC visual novel game on vocabulary learning}.
\newblock \bibinfo{journal}{\emph{Computer Assisted Language Learning}} \bibinfo{volume}{36}, \bibinfo{number}{3} (\bibinfo{year}{2021}), \bibinfo{pages}{312--345}.
\newblock
\href{https://doi.org/10.1080/09588221.2021.1928226}{doi:\nolinkurl{10.1080/09588221.2021.1928226}}


\bibitem[Marsh et~al\mbox{.}(2011)]%
        {marsh2011fun}
\bibfield{author}{\bibinfo{person}{Tim Marsh}, \bibinfo{person}{Li~Zhiqiang Nickole}, \bibinfo{person}{Eric Klopfer}, \bibinfo{person}{Chuang Xuejin}, \bibinfo{person}{Scot Osterweil}, {and} \bibinfo{person}{Jason Haas}.} \bibinfo{year}{2011}\natexlab{}.
\newblock \showarticletitle{Fun and learning: Blending design and development dimensions in serious games through narrative and characters}.
\newblock \bibinfo{journal}{\emph{Serious games and edutainment applications}} (\bibinfo{year}{2011}), \bibinfo{pages}{273--288}.
\newblock


\bibitem[Melcer et~al\mbox{.}(2020a)]%
        {melcer2020getting}
\bibfield{author}{\bibinfo{person}{Edward~F Melcer}, \bibinfo{person}{Katelyn~M Grasse}, \bibinfo{person}{James Ryan}, \bibinfo{person}{Nick Junius}, \bibinfo{person}{Max Kreminski}, \bibinfo{person}{Dietrich Squinkifer}, \bibinfo{person}{Brent Hill}, {and} \bibinfo{person}{Noah Wardrip-Fruin}.} \bibinfo{year}{2020}\natexlab{a}.
\newblock \showarticletitle{Getting academical: a choice-based interactive storytelling game for teaching responsible conduct of research}. In \bibinfo{booktitle}{\emph{Proceedings of the 15th International Conference on the Foundations of Digital Games}}. \bibinfo{pages}{1--12}.
\newblock


\bibitem[Melcer et~al\mbox{.}(2020b)]%
        {melcer2020teaching}
\bibfield{author}{\bibinfo{person}{Edward~F Melcer}, \bibinfo{person}{James Ryan}, \bibinfo{person}{Nick Junius}, \bibinfo{person}{Max Kreminski}, \bibinfo{person}{Dietrich Squinkifer}, \bibinfo{person}{Brent Hill}, {and} \bibinfo{person}{Noah Wardrip-Fruin}.} \bibinfo{year}{2020}\natexlab{b}.
\newblock \showarticletitle{Teaching responsible conduct of research through an interactive storytelling game}. In \bibinfo{booktitle}{\emph{Extended Abstracts of the 2020 CHI Conference on Human Factors in Computing Systems}}. \bibinfo{pages}{1--10}.
\newblock


\bibitem[Millard(2023)]%
        {millard2023strange}
\bibfield{author}{\bibinfo{person}{David~E Millard}.} \bibinfo{year}{2023}\natexlab{}.
\newblock \showarticletitle{Strange Patterns: Structure and Post-structure in Interactive Digital Narratives}.
\newblock \bibinfo{journal}{\emph{The Authoring Problem: Challenges in Supporting Authoring for Interactive Digital Narratives}} (\bibinfo{year}{2023}), \bibinfo{pages}{147--169}.
\newblock


\bibitem[Miller(2019)]%
        {miller2019digital}
\bibfield{author}{\bibinfo{person}{Carolyn~Handler Miller}.} \bibinfo{year}{2019}\natexlab{}.
\newblock \bibinfo{booktitle}{\emph{Digital Storytelling 4E: A creator’s guide to interactive entertainment}}.
\newblock \bibinfo{publisher}{CRC Press}, \bibinfo{address}{Boca Raton, FL}. 34 pages.
\newblock


\bibitem[Muntean et~al\mbox{.}(2018)]%
        {Muntean2018}
\bibfield{author}{\bibinfo{person}{Cristina~Hava Muntean}, \bibinfo{person}{Nour~El Mawas}, \bibinfo{person}{Michael Bradford}, {and} \bibinfo{person}{Pramod Pathak}.} \bibinfo{year}{2018}\natexlab{}.
\newblock \showarticletitle{Investigating the Impact of an Immersive Computer-Based Math Game on the Learning Process of Undergraduate Students}. In \bibinfo{booktitle}{\emph{2018 IEEE Frontiers in Education Conference (FIE)}}. \bibinfo{publisher}{IEEE}, \bibinfo{pages}{1--8}.
\newblock


\bibitem[Niedenthal(2009)]%
        {niedenthal2009what}
\bibfield{author}{\bibinfo{person}{Simon Niedenthal}.} \bibinfo{year}{2009}\natexlab{}.
\newblock \showarticletitle{What we talk about when we talk about game aesthetics}. In \bibinfo{booktitle}{\emph{Proceedings of the DiGRA 2009 Conference}}. \bibinfo{address}{London}.
\newblock


\bibitem[Nugroho et~al\mbox{.}(2018)]%
        {Nugroho2018}
\bibfield{author}{\bibinfo{person}{S.~M.~S. Nugroho}, \bibinfo{person}{A.~S. Utama}, \bibinfo{person}{M. Hariadi}, \bibinfo{person}{U.~L. Yuhana}, {and} \bibinfo{person}{M.~H. Purnomo}.} \bibinfo{year}{2018}\natexlab{}.
\newblock \showarticletitle{HEIRDOM: Multiple Ending Scenario Game For Mathematics Learning Using Rule-Based System}. In \bibinfo{booktitle}{\emph{2018 International Conference on Computer Engineering, Network and Intelligent Multimedia (CENIM)}}. \bibinfo{pages}{192--197}.
\newblock


\bibitem[Oliver et~al\mbox{.}(2016)]%
        {oliver2016video}
\bibfield{author}{\bibinfo{person}{Mary~Beth Oliver}, \bibinfo{person}{Nicholas~David Bowman}, \bibinfo{person}{Julia~K Woolley}, \bibinfo{person}{Ryan Rogers}, \bibinfo{person}{Brett~I Sherrick}, {and} \bibinfo{person}{Mun-Young Chung}.} \bibinfo{year}{2016}\natexlab{}.
\newblock \showarticletitle{Video games as meaningful entertainment experiences.}
\newblock \bibinfo{journal}{\emph{Psychology of Popular Media Culture}} \bibinfo{volume}{5}, \bibinfo{number}{4} (\bibinfo{year}{2016}), \bibinfo{pages}{390}.
\newblock


\bibitem[Putri et~al\mbox{.}(2021)]%
        {Putri2021}
\bibfield{author}{\bibinfo{person}{A. Putri}, \bibinfo{person}{N.~Z. Aida}, \bibinfo{person}{R.~A. Putri}, \bibinfo{person}{T.~S. Arrahmah}, {and} \bibinfo{person}{D. Kusrini}.} \bibinfo{year}{2021}\natexlab{}.
\newblock \showarticletitle{A3! visual novel game as an audio-visual media that motivates Japanese language learning}. In \bibinfo{booktitle}{\emph{Advances in Social Science, Education and Humanities Research}}.
\newblock
\href{https://doi.org/10.2991/assehr.k.211119.011}{doi:\nolinkurl{10.2991/assehr.k.211119.011}}


\bibitem[Rowe et~al\mbox{.}(2010)]%
        {rowe2010framework}
\bibfield{author}{\bibinfo{person}{Jonathan~P Rowe}, \bibinfo{person}{Lucy~R Shores}, \bibinfo{person}{Bradford~W Mott}, {and} \bibinfo{person}{James~C Lester}.} \bibinfo{year}{2010}\natexlab{}.
\newblock \showarticletitle{A framework for narrative adaptation in interactive story-based learning environments}. In \bibinfo{booktitle}{\emph{Proceedings of the intelligent narrative technologies III workshop}}. \bibinfo{pages}{1--8}.
\newblock


\bibitem[Ryan et~al\mbox{.}(2006)]%
        {ryan2006motivational}
\bibfield{author}{\bibinfo{person}{Richard~M Ryan}, \bibinfo{person}{C~Scott Rigby}, {and} \bibinfo{person}{Andrew Przybylski}.} \bibinfo{year}{2006}\natexlab{}.
\newblock \showarticletitle{The motivational pull of video games: A self-determination theory approach}.
\newblock \bibinfo{journal}{\emph{Motivation and Emotion}}  \bibinfo{volume}{30} (\bibinfo{year}{2006}), \bibinfo{pages}{344--360}.
\newblock


\bibitem[Salazar et~al\mbox{.}(2013)]%
        {Salazar2013}
\bibfield{author}{\bibinfo{person}{Francisco~Lepe Salazar}, \bibinfo{person}{Tatsuo Nakajima}, {and} \bibinfo{person}{Todorka Alexandrova}.} \bibinfo{year}{2013}\natexlab{}.
\newblock \showarticletitle{Visual Novels: A Methodology Guideline for Pervasive Educational Games that Favors Discernment}. In \bibinfo{booktitle}{\emph{International Conference on Grid and Pervasive Computing}}. \bibinfo{publisher}{Springer}, \bibinfo{pages}{234--243}.
\newblock


\bibitem[Salter(2016)]%
        {salter2016}
\bibfield{author}{\bibinfo{person}{Anastasia Salter}.} \bibinfo{year}{2016}\natexlab{}.
\newblock \showarticletitle{Playing at empathy: Representing and experiencing emotional growth through Twine games}. In \bibinfo{booktitle}{\emph{Proc. International Conference on Serious Games and Applications for Health}}. \bibinfo{publisher}{IEEE}, \bibinfo{pages}{1--8}.
\newblock


\bibitem[Schmaling and Blume(2009)]%
        {schmaling2009}
\bibfield{author}{\bibinfo{person}{Karen~B Schmaling} {and} \bibinfo{person}{Arthur~W Blume}.} \bibinfo{year}{2009}\natexlab{}.
\newblock \showarticletitle{Ethics instruction increases graduate students’ responsible conduct of research knowledge but not moral reasoning}.
\newblock \bibinfo{journal}{\emph{Accountability in Research}} \bibinfo{volume}{16}, \bibinfo{number}{5} (\bibinfo{year}{2009}), \bibinfo{pages}{268--283}.
\newblock


\bibitem[Seiler et~al\mbox{.}(2011)]%
        {seiler2011}
\bibfield{author}{\bibinfo{person}{Stephanie~N Seiler}, \bibinfo{person}{Bradley~J Brummel}, \bibinfo{person}{Kerri~L Anderson}, \bibinfo{person}{Kyoung~Jin Kim}, \bibinfo{person}{Serena Wee}, \bibinfo{person}{CK Gunsalus}, {and} \bibinfo{person}{Michael~C Loui}.} \bibinfo{year}{2011}\natexlab{}.
\newblock \showarticletitle{Outcomes assessment of role-play scenarios for teaching responsible conduct of research}.
\newblock \bibinfo{journal}{\emph{Accountability in Research}} \bibinfo{volume}{18}, \bibinfo{number}{4} (\bibinfo{year}{2011}), \bibinfo{pages}{217--246}.
\newblock


\bibitem[Short(2019)]%
        {short2019storylets}
\bibfield{author}{\bibinfo{person}{Emily Short}.} \bibinfo{year}{2019}\natexlab{}.
\newblock \bibinfo{title}{Storylets: You Want Them}.
\newblock
\urldef\tempurl%
\url{https://emshort.blog/2019/11/29/storylets-you-want-them/}
\showURL{%
\tempurl}


\bibitem[Simons(2007)]%
        {simons2007narrative}
\bibfield{author}{\bibinfo{person}{Jan Simons}.} \bibinfo{year}{2007}\natexlab{}.
\newblock \showarticletitle{Narrative, games, and theory}.
\newblock \bibinfo{journal}{\emph{Game Studies}} \bibinfo{volume}{7}, \bibinfo{number}{1} (\bibinfo{year}{2007}).
\newblock
\urldef\tempurl%
\url{http://www.gamestudies.org/07010701/articles/simons}
\showURL{%
\tempurl}


\bibitem[Sullivan and Critten(2014)]%
        {Sullivan2014}
\bibfield{author}{\bibinfo{person}{Dean Sullivan} {and} \bibinfo{person}{Jessica Critten}.} \bibinfo{year}{2014}\natexlab{}.
\newblock \showarticletitle{Adventures in Research: Creating a Video Game Textbook for an Information Literacy Course}.
\newblock \bibinfo{journal}{\emph{College \& Research Libraries News}} \bibinfo{volume}{75}, \bibinfo{number}{10} (\bibinfo{year}{2014}), \bibinfo{pages}{570--573}.
\newblock
\href{https://doi.org/10.5860/crln.75.10.9215}{doi:\nolinkurl{10.5860/crln.75.10.9215}}


\bibitem[Wardrip and Jhala(2014)]%
        {wardrip2014towards}
\bibfield{author}{\bibinfo{person}{Peter Mawhorter Michael Mateas~Noah Wardrip} {and} \bibinfo{person}{Fruin~Arnav Jhala}.} \bibinfo{year}{2014}\natexlab{}.
\newblock \showarticletitle{Towards a theory of choice poetics}. In \bibinfo{booktitle}{\emph{In Proceedings of the 9th International Conference on the Foundations of Digital Games}}.
\newblock


\bibitem[Wardrip-Fruin et~al\mbox{.}(2009)]%
        {wardripfruin2009agency}
\bibfield{author}{\bibinfo{person}{Noah Wardrip-Fruin}, \bibinfo{person}{Michael Mateas}, \bibinfo{person}{Steven Dow}, {and} \bibinfo{person}{Serdar Sali}.} \bibinfo{year}{2009}\natexlab{}.
\newblock \showarticletitle{Agency reconsidered}. In \bibinfo{booktitle}{\emph{Proceedings of DiGRA 2009 Conference: Breaking New Ground: Innovation in Games, Play, Practice and Theory}}.
\newblock


\bibitem[Wei et~al\mbox{.}(2010)]%
        {wei2010time}
\bibfield{author}{\bibinfo{person}{Huaxin Wei}, \bibinfo{person}{Jim Bizzocchi}, {and} \bibinfo{person}{Tom Calvert}.} \bibinfo{year}{2010}\natexlab{}.
\newblock \showarticletitle{Time and space in digital game storytelling}.
\newblock \bibinfo{journal}{\emph{International Journal of Computer Games Technology}} \bibinfo{volume}{2010}, \bibinfo{number}{1} (\bibinfo{year}{2010}), \bibinfo{pages}{897217}.
\newblock


\bibitem[Whitbeck(2001)]%
        {whitbeck2001}
\bibfield{author}{\bibinfo{person}{Caroline Whitbeck}.} \bibinfo{year}{2001}\natexlab{}.
\newblock \showarticletitle{Group mentoring to foster the responsible conduct of research}.
\newblock \bibinfo{journal}{\emph{Science and Engineering Ethics}} \bibinfo{volume}{7}, \bibinfo{number}{4} (\bibinfo{year}{2001}), \bibinfo{pages}{541--558}.
\newblock


\bibitem[Yin et~al\mbox{.}(2012)]%
        {Yin2012}
\bibfield{author}{\bibinfo{person}{Langxuan Yin}, \bibinfo{person}{Lazlo Ring}, {and} \bibinfo{person}{Timothy Bickmore}.} \bibinfo{year}{2012}\natexlab{}.
\newblock \showarticletitle{Using an Interactive Visual Novel to Promote Patient Empowerment through Engagement}. In \bibinfo{booktitle}{\emph{Proceedings of the International Conference on the Foundations of Digital Games}}. \bibinfo{pages}{41--48}.
\newblock


\end{thebibliography}

\appendix
\section{Codes with example quotes from the participants}
\label{code:appendix}
\begin{table*}[hb]
  %\caption{Codes with example quotes from the participants}
  \label{tab:example-quotes}
  \Description{Codes with example quotes from the participants. Themes are 'Motivational Factors,' 'Game elements,' and 'Perception.' Under motivational factors, there are Competence - “I most enjoyed how simple it was to progress through the game.” P44 (2.0), Autonomy “Being able to make my own choices was enjoyable.” P24 (1.0), Relatedness “I have had a similar experience and my paper was rejected for not following IRB guidelines.” P34 (2.0). Under Game elements, there are Aesthetics “Illustrations were great too; the facial expression hurt.” P50 (2.0), Narrative “The story was fun to follow. I liked how different options resulted in different outcomes” P15 (1.0), and under perception, there are Simplicity “The choices were quite limited. The game was too short, and you couldn’t see any consequences from your choices.” P21 (1.0) and Loop “Sometimes it felt like I was going in a loop and I was trying to figure out how to progress forward.” P49 (2.0).}
  \centering
  \begin{tabular}{p{0.15\textwidth}p{0.12\textwidth}p{0.7\textwidth}}
    \toprule
    \textbf{Theme} & \textbf{Sub-themes} & \textbf{Example quotes} \\
    \midrule
    \textbf{Motivational factors} 
    & Competence  & “I most enjoyed how simple it was to progress through the game.” P44 (2.0) \\
    & Autonomy    & “Being able to make my own choices was enjoyable.” P24 (1.0)\\
    & Relatedness & “I have had a similar experience and my paper was rejected for not following IRB guidelines.” P34 (2.0) \\
    \textbf{Game elements}  
    & Aesthetics  & “Illustrations were great too; the facial expression hurt.” P50 (2.0) \\
    & Narrative   & “The story was fun to follow. I liked how different options resulted in different outcomes.” P15 (1.0) \\
    \textbf{Perception}     
    & Simplicity  & “The choices were quite limited. The game was too short, and you couldn't see any consequences from your choices.” P21 (1.0)\\
    & Loop        & “Sometimes it felt like I was going in a loop and I was trying to figure out how to progress forward.” P49 (2.0) \\
    \bottomrule
  \end{tabular}
\end{table*}

\end{document}